# Sidestepping the Triangulation Problem in Bayesian Net Computations

(Nevin) Lianwen Zhang and David Poole
Computer Science Dept.
University of British Columbia
Vancover, B. C. V6T 1Z2
CANADA
e-mail: lzhang@cs.ubc.ca poole@cs.ubc.ca

## Abstract

This paper presents a new approach for computing posterior probabilities in Bayesian nets, which sidesteps the triangulation problem. The current state of art is the clique tree propagation approach. When the underlying graph of a Bayesian net is triangulated, this approach arranges its cliques into a tree and computes posterior probabilities by appropriately passing around messages in that tree. The computation in each clique is simply direct marginalization. When the underlying graph is not triangulated, one has to first triangulated it by adding edges. Referred to as the triangulation problem, the problem of finding an optimal or even a "good" triangulation proves to be difficult. In this paper, we propose to first decompose a Bayesian net into smaller components by making use of Tarjan's algorithm for decomposing an undirected graph at all its minimal complete separators. Then, the components are arranged into a tree and posterior probabilities are computed by appropriately passing around messages in that tree. The computation in each component is carried out by repeating the whole procedure from the beginning. Thus the triangulation problem is sidestepped.

## 1 INTRODUCTION

There has been in recent years extensive research in computing marginal and posterior marginal probabilities in Bayesian nets. Largely due to the work of Pearl (1988), Lauritzen and Spiegelhalter (1988), Shafer and Shenoy (1988), and Jensen *et al.* (1990), an efficient algorithm called clique tree propagation has come into being. When the underlying graphs are triangulated, the algorithm arranges the cliques of the underlying graphs into join trees and obtains the wanted marginals and posteriors by passing messages around in those trees. The computation in each clique is simply direct marginalization. When the underlying graphs are not triangulated, one needs to triangulate them first (Lauritzen and Spiegelhalter 1988), or appeal to the technique of conditioning on loop cutsets (Pearl 1988). A similar preprocessing step is also required in the goal directed approach of shacter *et al.* (1990). In this paper, we shall refer to the problem of finding an optimal or a "good" triangulation for an untriangulated graph as the triangulation problem.

It is NP-hard to find either optimal loop cutsets (Stillman 1990) or optimal triangulations (Yannakakis 1981). Although heuristics are available, Stillman (1990) has "demonstrated that no heuristic algorithm .... can be guaranteed to produce loop cutsets within a constant difference from optimal". We conjecture that the same thing is true for triangulation. Thus, it is interesting to investigate the possibility of approaches that sidestep the triangulation problem.

Poole and Neufeld (1991) describes an interesting implementation of Bayesian nets in Prolog. The implementation encodes conditional probabilities as Prolog goal reduction rules and it organizes reasoning by using Bayes' Theorem and causality relationships among the variables. The need for triangulation is avoided.

In this paper, we describe a new approach for computing posterior probabilities in Bayesian nets, which also sidesteps the triangulation problem. What differentiates our approach from Poole and Neufeld's approach is that the former makes full use of decompositions, while the latter does not at all. The cornerstone of our approach is Tarjan's algorithm for decomposing undirected graphs.

An undirected graph may contain many complete separators even if it not triangulated. Tarjan (1985) presents a $O(nm)$ time algorithm for decomposing an undirected graph at all its minimal complete separators, where $n$ and $m$ are the numbers of vertices and edges of in the graph respectively. It is easy to conceive a procedure that decomposes a Bayesian net into as many components as possible by using Tarjan's algorithm on the moral graph.

Our approach can be viewed as an advancement of the clique tree propagation approach. Like the clique tree approach, this approach first arranges the components of a Bayesian net into a tree, and then computes posterior probabilities by appropriately passing information around in that tree. Unlike the clique tree approach, the computation in each component is not carried out by direct marginalization. Rather it is carried by repeating the whole procedure from the beginning.

The organization of the paper is as follows. Section 2 starts the paper by reviewing some graph theory terminologies and some basic concepts pertaining to Bayesian nets. Section 3 introduces the concept of decomposition in terms of semi-Bayesian nets. Section 4 shows how decomposition leads to the parallel reduction technique. Section 5 introduces serial reduction – the basic means for passing messages from one component to another, and section 6 constructs the component tree – the basic means for organizing message passing. The algorithm is given in section 7, together with an example illustrating how it works. The paper concludes at section 8.

## 2  PREREQUISITE

Let us begin by reviewing some graph theory terminologies. An (undirected) *graph* $G = (V, E)$ consists of a set $V$ of *vertices* and a set $E$ of *edges*, which are (unordered) pairs of vertices. A *path* is a sequence of vertices in which every pair of consecutive vertices is an edge. A *loop* is a path where no vertex appear twice except the first one and the last one, which are the same. A *chordless* loop is a loop in which no pair of non-consecutive vertices is an edge. A *genuine loop* is a chordless loop of length greater than three. A *triangulated* graph is one without genuine loops.

A graph is *connected* if there is a path between any two vertices. If a graph $G$ is disconnected, a *connected component* of $G$ consists of a subset of vertices in which vertices are connected to each other but not to vertices outside the subset, and of all the edges that are composed of vertices in the subset.

A *separator* of a graph is a subset of its vertices whose deletion from the graph will leave it disconnected. Two vertices is *separated* by a separator if every path connecting them contains at least one vertex in the separator. A separator is *minimal* if none of it proper subsets are separators. A subset of vertices is *complete* if every pair of its elements is an edge. Separators can be complete.

*Cliques* are maximal complete subsets of vertices. When the set of all the vertices is complete, we say that the graph is complete.

A *directed graph* $G = (V, A)$ consists of a set $V$ of *vertices* and a set $A$ of *arcs*, which are ordered pairs of vertices. If there is an arc from vertex $v_1$ to vertex $v_2$, then $v_1$ is a *parent* of $v_2$ and $v_2$ is a *child* of $v_1$. Vertices with no children are *leaves* and vertices with no parents are *roots*. The set of all the parents of a vertex $v$ will be notated by $\pi(v)$. A *directed cycle* is a sequence of vertices in which every vertex is a parent of the vertex after it and the last vertex is a parent of the first vertex. An *acyclic directed* graph is one in which there are no directed cycles.

To *marry* all the parents of a vertex in a directed graph is to add an (undirected) edge between each pair of its parents. The *moral graph* $m(G)$ of a directed graph $G$ is an undirected graph obtained from $G$ by marrying all the parents of each vertex respectively and ignoring all the directions on the arcs. A directed graph *is connected* if its moral graph is.

Let us now review a few concepts pertaining to Bayesian nets. A *Bayesian net* $\mathcal{N}$ is a triplet $(V, A, \mathcal{P})$, where

1. $V$ is a set of variables.
2. $A$ is a set of arcs, which together with $V$ constitutes a directed acyclic graph $G = (V, A)$.
3. $\mathcal{P} = \{P(v|\pi(v)) : v \in V\}$, i.e $\mathcal{P}$ is the set the conditional probabilities of the all variables given their respective parents [1].

Variables in a Bayesian net will also be referred to as vertices and nodes when the emphasis is on the underlying graph.

The *prior joint probability* $P_\mathcal{N}$ of a Bayesian net $\mathcal{N}$ is defined by

$$P_\mathcal{N}(V) = \prod_{v \in V} P(v|\pi(v)). \qquad (1)$$

Observations may be made about variables. Let $Y \subseteq V$ be the set of variables observed and $Y_0$ be the corresponding set of values. Let $X \subseteq V$ be the set of variables of interest. The *posterior probability* $P_\mathcal{N}(X|Y = Y_0)$ of $X$ in a Bayesian net $\mathcal{N}$ given observations $Y = Y_0$ is obtained by first conditioning $P_\mathcal{N}$ on $Y = Y_0$ and marginalizing the resulting joint probability onto $X$ [2]. The problem of concern to this paper is how to compute $P(X|Y = Y_0)$?

Developed by Lauritzen and Spiegehalter (1988), Shafer and Shenoy (1988), Jensen *et al* (1990), the clique tree propagation approach consists of three steps: (1) Triangulate the moral graph of $G$; (2) arrange the cliques into a join tree, and (3) properly pass messages around in the tree.

Consider the Bayesian net **net1** in Figure 1 (1). The following prior probability and conditional probabilities are given as part of the specification: $P(c)$, $P(a|c)$, $P(e|a)$, $P(f|e)$, $P(g|f)$, $P(b|a,g)$, $P(h|b)$, and $P(d|c, h)$. The moral graph of its underlying graph is shown 1 (2). The dashed edges are added for triangulation. The corresponding join tree is shown in Figure 1 (3). One can obtain $P_{net1}(d|e = e_0)$ by passing messages in the way as indicated by the arrows. See the cited papers for details about the contents of the messages and how are they actually passed in the join tree.

---

[1]Note that when $v$ is a root $\pi(v)$ is empty. In such a case, the expression $P(v|\pi(v))$ simply stands for the prior probability of $v$.

[2]That is to sum out all the variables not in $X$.

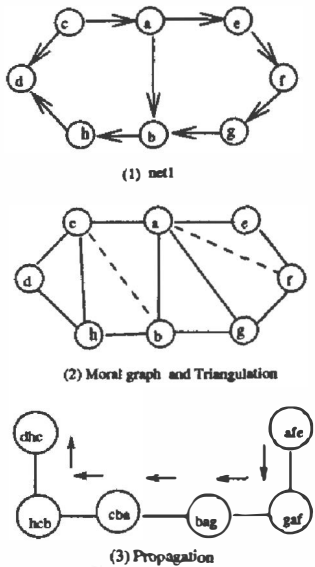

Figure 1: Bayesian net, triangulation and message passing.

Let $X$ be the set of variables of interest. Let $Y$ be the set of observed variables, and $Y_0$ be the corresponding set of observed values. We are concerned with queries of the form $P_\mathcal{N}(X|Y = Y_0)$. Noticing that $P_\mathcal{N}(X|Y = Y_0)$ can be obtained normalizing the marginal probability $P_\mathcal{N}(X, Y = Y_0)$, we shall concern ourselves with queries of the form $P_\mathcal{N}(X, Y = Y_0)$ in this paper.

## 3  DECOMPOSITION

In this section, we shall introduce the concept of decomposition in terms semi-Bayesian nets.

A *semi-Bayesian net* $\mathcal{N}$ is a quadruplet $\mathcal{N} = (V, A, R, \mathcal{P})$, where

1. $V$ is a set of variables.
2. $A$ is the set of arcs, which together with $V$ constitutes a directed acyclic graph $G = (V, A)$.
3. $R$ is a set of roots of $G$.
4. $\mathcal{P} = \{P(v|\pi(v)) : v \in (V - R)\}$.

In words, semi-Bayesian net are Bayesian nets with *unspecified roots*, i.e with roots whose prior probabilities are not given. Thus, Bayesian nets are semi-Bayesian nets whose set of unspecified roots is empty.

In a semi-Bayesian net $\mathcal{N} = (V, A, R, \mathcal{P})$, the $R$ is the set of roots of the directed graph $(V, A)$ whose prior probabilities are not given in $\mathcal{P}$. So, we call $R$ the *set of unspecified roots of* $(V, A, \mathcal{P})$.

The *prior joint potential* $P_\mathcal{N}$ a semi-Bayesian net $\mathcal{N} = (V, G, R, \mathcal{P})$ is defined by

$$P_\mathcal{N}(V) = \prod_{v \in V - R} P(v|\pi(v)). \quad (2)$$

And for any subset $X$ of $V$, the *(marginal) potential* $P_\mathcal{N}(X)$ of $X$ in $\mathcal{N}$ is obtained by marginalizing $P_\mathcal{N}(V)$ onto $X$, i.e by summing out all the variables in $V - X$.

If $\mathcal{N}$ is a Bayesian net, then its prior joint potential and marginal potentials are exactly the same as its prior joint probability and marginal probabilities defined in the previous section.

Let us now turn to decomposition of undirected graphs. An undirected graph $G = (V, E)$ *decomposes* into $G_1 = (V_1, E_1)$ and $G_2 = (V_2, E_2)$, or simply into $V_1$ and $V_2$ if:

1. $V_1$ and $V_2$ are proper subsets of $V$ and $V_1 \cup V_2 = V$,
2. $E_i$ ($i = 1, 2$) consists of all the edges in $E$ that lie completely within $V_i$ and $E_1 \cup E_2 = E$, and
3. $V_1 \cap V_2$ is a complete separator of $G$.

In such a case, we say that $G$ is *decomposable*.

There is a one-to-one correspondence between complete separators and decompositions. Given any complete separator $S$ of an undirected graph $G$, $G$ can be decomposed at $S$ into to $V_1$ and $V_2$ such that $V_1 \cap V_2 = S$. And if $G$ decomposes into $V_1$ and $V_2$, then $V_1 \cap V_2$ is a complete separator. Tarjan (1985) presents an $O(nm)$ time algorithm for decomposing an undirected graph at all its minimal complete separators, where $n$ and $m$ stands for the numbers of vertices and edges of $G$ respectively. Tarjan's algorithm is very important to our approach for computing posterior probabilities in Bayesian nets.

A semi-Bayesian net $\mathcal{N} = (V, A, R, \mathcal{P})$ is *decomposable* if the moral graph $m(G)$ of the underlying directed graph $G = (V, A)$ is decomposable. A decomposition $\{V_1, V_2\}$ of $m(G)$ induces *the decomposition of the semi-Bayesian net $\mathcal{N}$ into $\mathcal{N}_1 = (V_1, A_1, R_1, \mathcal{P}_1)$ and $\mathcal{N}_2 = (V_2, A_2, R_2, \mathcal{P}_2)$*, where

- $\mathcal{P}_1$ is obtained from $\mathcal{P}$ by removing all the items that do not involve any variables in $V_1 - V_2$ and $\mathcal{P}_2 = \mathcal{P} - \mathcal{P}_1$.
- And for $i \in \{1, 2\}$,
  - $A_i$ is obtained $A$ by removing any arcs whose two vertices do not simultaneously appear in any item of $\mathcal{P}_i$, and

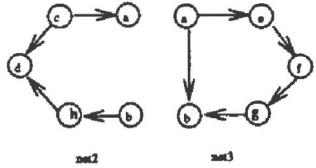

Figure 2: A decomposition of **net1**.

- $R_i$ is the set of unspecified roots of $(V_i, A_i, \mathcal{P}_i)$.

Consider the Bayesian net **net1** in Figure 1 (a). The set $\{a, b\}$ is a minimal complete separator. The moral graph of the directed graph decomposes into $\{a, b, c, d, h\}$ and $\{a, b, e, f, g\}$. This induces the decomposition of **net1** into semi-Bayesian nets **net2** and **net3** as shown in Figure 2. In **net2**, $P(c)$, $P(a|c)$, $P(h|b)$, and $P(d|b, c)$ are inherited from **net1**. There is no arc from $a$ to $b$ because they do not simultaneously appear in any of those (conditional) probabilities of **net2**. The root $b$ of **net2** is unspecified because there is no $P(b)$ in **net2**. The conditional probabilities in **net3** are $P(e|a)$, $P(f|e)$, $P(g|f)$, and $P(b|a, g)$. The root $a$ is unspecified because there is no $P(a)$.

A semi-Bayesian net is *simple* if it contains only one leaf node and all other nodes are parents of this leaf node. In any semi-Bayesian net that is not simple, there is always a leaf node, say $l$. The set $\pi(l)$ of the parents of $l$ is a complete separator, which separates $l$ from all the nodes not in $\pi(l) \cup \{l\}$. So, there always a minimal complete separator $S$ that separates $l$ from all the nodes not in $S \cup \{l\}$. Thus we have

**Proposition 1** *If a semi-Bayesian net is not simple, then it is decomposable.*

Remark: We do not want to decompose a Bayesian net at a separator that is not complete for two reasons. First, it increases the complexity of the problem itself, and second the number of separators that are not complete may be too large.

## 4 PARALLEL REDUCTION

In this section, we shall show how the concept of decomposition can be used in computing marginal potentials in semi-Bayesian nets (marginal probabilities in Bayesian nets). First of all, we have the following theorem.

**Theorem 2** *Suppose a semi-Bayesian net $\mathcal{N}$ decomposes into $\mathcal{N}_1$ and $\mathcal{N}_2$ at a minimal complete separator $S$ (of the moral graph of its underlying directed graph). Let $X$ be a subset of variables of $\mathcal{N}_1$ which do not appear in $S$, let $Y$ bet a subset of variables of $\mathcal{N}_2$ which do not appear in $S$, and let $Z$ be a subset of $S$. Then*

$$P_\mathcal{N}(X, Z, Y) = \sum_{S-Z} P_{\mathcal{N}_1}(X, S) P_{\mathcal{N}_2}(S, Y). \quad (3)$$

The theorem follows from the definition decompositions of semi-Bayesian nets and the distributivity of multiplication w.r.t summation. Instead of giving the detailed proof, we shall provide an illustrating example.

Suppose we want to compute $P_{net1}(d, a, e)$ in the Bayesian net **net1** shown Figure 1 (1). By definition, we have

$$\begin{aligned}
P_{net1}&(d, a, e) \\
&= \sum_{c, b, f, g, h} P(c) P(a|c) P(e|a) P(f|e) P(g|f) \\
&\qquad\qquad P(b|a, g) P(h|b) P(d|c, b) \\
&= \sum_b \{\sum_{c, h} P(c) P(a|c) P(h|b) P(d|c, b)\} \\
&\qquad\qquad \{\sum_{f, g} P(e|a) P(f|e) P(g|f) P(b|a, g)\} \quad (4)\\
&= \sum_b P_{net2}(d, a, b) P_{net3}(a, b, e). \quad (5)
\end{aligned}$$

Equation (4) is true because of the distributivity of multiplication w.r.t summation, and equation (5) immediately follows from the definition of marginal potentials in semi-Bayesian nets.

Suppose we want to compute the marginal potential $P_\mathcal{N}(X, Y = Y_0)$ in a semi-Bayesian net $\mathcal{N}$. And suppose $\mathcal{N}$ decomposes into $\mathcal{N}_1$ and $\mathcal{N}_2$ at a minimal complete separator $S$ of the moral graph of its underlying directed graph. Let $X_1$ be the set of variables in $X$ and in $\mathcal{N}_1$ but not in $S$, $X_2$ be the set of variables in $X$ and in $\mathcal{N}_2$ but not in $S$, and $X_S = X \cap S$. The sets $Y_1$, $Y_2$ and $Y_S$ are defined from $Y$ in the same way. Let $Y_{-1} = Y_S \cup Y_2$ and $Y_{-2} = Y_S \cup Y_1$. Let $(Y_{-1})_0$ and $(Y_{-2})_0$ be the corresponding sets of values of $Y_{-1}$ and $Y_{-2}$.

*The query induced in $\mathcal{N}_1$ by $P_\mathcal{N}(X, Y = Y_0)$ is $P_{\mathcal{N}_1}(X_1, S - Y_S, Y_{-2} = (Y_{-2})_0)$, and the query induced in $\mathcal{N}_2$ is $P_{\mathcal{N}_2}(X_2, S - Y_S, Y_{-1} = (Y_{-1})_0)$.* According to Theorem 2, we have

$$\begin{aligned}
P_\mathcal{N}&(X, Y = Y_0) \\
&= \sum_{S - X_S - Y_S} P_{\mathcal{N}_1}(X_1, S - Y_S, Y_{-2} = (Y_{-2})_0) \\
&\qquad\qquad P_{\mathcal{N}_2}(X_2, S - Y_S, Y_{-1} = (Y_{-1})_0) \quad (6)
\end{aligned}$$

Thus to compute $P_\mathcal{N}(X, Y = Y_0)$, we can first compute $P_{\mathcal{N}_1}(X_1, S - Y_S, Y_{-2} = (Y_{-2})_0)$ and $P_{\mathcal{N}_2}(X_2, S - Y_S, Y_{-1} = (Y_{-1})_0)$ (possibly in parallel), and then combine the results by equation (6). This leads to the technique of parallel reduction.

Given a semi-Bayesian net $\mathcal{N}$, the following procedure computes $P_\mathcal{N}(X, Y = Y_0)$:

**Parallel-Reduction**

> **If** $\mathcal{N}$ is simple, then compute $P_\mathcal{N}(X, Y = Y_0)$ directly by marginalization,
> **else** find a minimal complete separator $S$ of the moral graph of the underlying directed graph of $\mathcal{N}$, decompose $\mathcal{N}$ at $S$ into $\mathcal{N}_1$ and $\mathcal{N}_2$.
> 1. Repeat the procedure to compute the induced query $P_{\mathcal{N}_1}(X_1, S - Y_S, Y_{-2} = (Y_{-2})_0)$,
> 2. Repeat the procedure to compute the induce query $P_{\mathcal{N}_2}(X_2, S - Y_S, Y_{-1} = (Y_{-1})_0)$,
> 3. Combine the answers to the two induced queries by using equation (6).

The procedure is termed *parallel reduction* because line 1 and line 2 can be executed strictly in parallel.

Because of Proposition 1, the procedure Parallel Reduction is able to compute marginal potentials in semi-Bayesian nets without triangulating the underlying graphs. But the algorithm can be very inefficient.

The main purpose of this paper is to describe another algorithm called component tree propagation, which we hope is as efficient as the clique tree propagation approach based on an optimal triangulation. The algorithm computes posterior probabilities in a Bayesian net as follows: first the Bayesian net is decomposed into components, the components are arranged into a tree, and then posterior probabilities are obtained by appropriately passing messages around in that tree. In the next section, we shall introduce the basic means for passing messages from one component to another — the serial reduction technique. In the section after, we shall present the basic means for controlling message passing — component trees.

## 5 SERIAL REDUCTION

Suppose a semi-Bayesian net $\mathcal{N}$ decomposes into two components. A query in $\mathcal{N}$ reduces into two subqueries, one in each of the components. The parallel reduction procedure first computes both of the two subqueries (possibly in parallel), and then use an additional formula to combine the answers to get the answer to the original query. In contrast, the serial reduction procedure will first compute only one of two subqueries. The answer is then send to the other subquery, which is thereby updated. The answer to the updated subquery is the same as the answer to the original query.

This section is devoted serial reduction. Let us begin by introducing the concept of parametric semi-Bayesian nets.

A *parametric semi-Bayesian net* $\mathcal{N} = (V, A, R, \mathcal{P})$ is a semi-Bayesian net, except that some of its conditional probabilities contain *parameters*, i.e variables that are not members of $V$. Thus, semi-Bayesian nets are parametric semi-Bayesian nets that do not contain any parameters.

Suppose $\mathcal{N}$ is a semi-Bayesian net with parameters $W$. And suppose we want to compute the potential of $(X, Y = Y_0)$. The answer will be a function of the parameters $W$ as well as of $X$. So, we shall write the query as $P_\mathcal{N}(X, Y = Y_0 : W)$, where the column mark separates parameters from variables in the semi-Bayesian net.

Given a query $P_\mathcal{N}(X, Y = Y_0 : W)$ in a parametric semi-Bayesian net $\mathcal{N}$, a node is *laden* if it is a leaf of $\mathcal{N}$ and it is in the set $Y$ (i.e it is observed).

Suppose $\mathcal{N}$ decompose into $\mathcal{N}_1$ and $\mathcal{N}_2$ at a minimal complete separator $S$. Let the sets $X_1$, $X_S$, $X_2$, $Y_1$, $Y_S$, $Y_2$, $Y_{-1}$, and $Y_{-2}$ are defined as before. Let $W_1$ and $W_2$ be the parameters of $\mathcal{N}_1$ and $\mathcal{N}_2$ respectively.

As in the case of parallel reduction, the query $Q$: $P_\mathcal{N}(X, Y = Y_0 : W)$ induces a subquery $Q_1$ in $\mathcal{N}_1$ and a subquery $Q_2$ in $\mathcal{N}_2$. The induced subquery $Q_1$ is $P_{\mathcal{N}_1}(X_1, S - Y_S, Y_{-2} = (Y_{-2})_0 : W_1)$, and the induced subquery $Q_2$ is $P_{\mathcal{N}_2}(X_2, S - Y_S, Y_{-1} = (Y_{-1})_0 : W_2)$.

The answer to the subquery $Q_2$ is a function $f_0(X_2, S - Y_S, W_2)$. We extend it to be a function $f(X_2, S, W_2)$ by setting

$$f(X_2, S, W_2) = \begin{cases} f_0(X_2, S - Y_S, W_2) & \text{if } Y_S = (Y_S)_0 \\ 0 & \text{otherwise} \end{cases}$$

To *append the answer* $f(X-2, S, W_2)$ *of* $Q_2$ *to* $\mathcal{N}_1$ is to

1. Introduce an auxiliary variable $v$ into $\mathcal{N}_1$, which has only two possible values 0 and 1,
2. Draw an arc to $v$ from each variable in $S$,
3. Set $P(v = 0|S) = f(X_2, S, W_2)$,

Let $\mathcal{N}_1'$ denote the resulting semi-Bayesian net. To *use the answer of $Q_2$ to update* the query $Q$ is to replace it by $Q'$: $P_{\mathcal{N}_1'}(X_1, X_S, Y_{-2} = (Y_{-2})_0, v = 0 : X_2, W)$.

Note that the auxiliary variable $v$ is a laden variable in the updated subquery. Also note that the the semi-Bayesian net $\mathcal{N}_1'$ contains the parameters $W \cup X_2$.

**Theorem 3** *Suppose $\mathcal{N}$ is semi-Bayesiant net with parameters $W$, which decomposes at a minimal complete separator $S$ into $\mathcal{N}_1$ and $\mathcal{N}_2$. Let all the symbols be as defined above. Then*

$$P_\mathcal{N}(X, Y = Y_0 : W) =$$
$$P_{\mathcal{N}_1'}(X_1, X_S, Y_{-2} = (Y_{-2})_0, v = 0 : X_2, W). \quad (7)$$

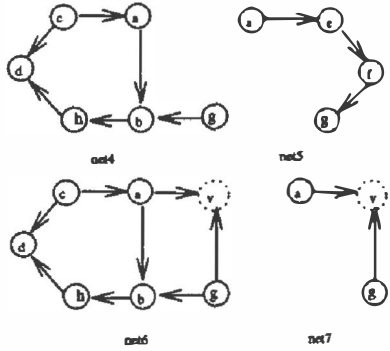

Figure 3: A serial reduction of net1.

The theorem says, to compute the query $P_\mathcal{N}(X, Y = Y_0 : W)$, we can first compute the induced subquery $P_{\mathcal{N}_2}(X_2, S - Y_S, Y_{-1} = (Y_{-1})_0 : W)$, use the answer to construct $\mathcal{N}'_1$, and then compute the updated query $P_{\mathcal{N}'_1}(X_1, X_S, Y_{-2} = (Y_{-2})_0 : X_2, W)$. This procedure is called *serial reduction*.

Again, instead of give the detailed proof of the theorem, we shall provide an illustrating example. Consider computing $P_{net1}(d, e)$ in the Bayesian net net1 shown in Figure 1 (1). The net decomposes into net4 and net5 (in Figure 3) at the minimal separator $\{a, g\}$. We first compute $P_{net5}(a, g, e)$, and append the result to net4. This gives us net6. Drawn in dotted cycle, $v$ is an auxiliary laden variable introduced. The conditional probability of $v = 0$ give $a$ and $g$ is set by $P(v = 0|a, g) = P_{net5}(a, g, e)$. Thus net6 contains the parameter $e$. The updated query is $P_{net6}(d, v = 0 : e)$. To see that $P_{net6}(d, v = 0 : e) = P_{net1}(d, e)$, we notice that net6 decomposes into net4 and net7, and that $P_{net7}(a, g, v = 0) = P(v = 0|a, g) = P_{net5}(a, g, e)$. So

$$P_{net6}(d, v = 0 : e)$$
$$= \sum_{a,g} P_{net4}(d, a, g) P_{net7}(a, g, v = 0)$$
$$= \sum_{a,g} P_{net4}(d, a, g) P_{net5}(a, g, e)$$
$$= P_{net1}(d, e).$$

## 6 COMPONENT TREES

Given a query $P_\mathcal{N}(X, Y = Y_0 : W)$ in a parametric semi-Bayesian net $\mathcal{N}$, a minimal complete separator (of the moral graph of the underlying directed graph of $\mathcal{N}$) is *trivial* if it is a subset of $\pi(l)$ for some laden node $l$ and its deletion from the moral graph only result in no more than two connected components.

Tarjan's algorithm decomposes an undirected graph at all its minimal complete separators. Based on this algorithm, one can easily conceive a procedure to find all the non-trivial minimal complete (NMC) separators of $\mathcal{N}$ and decompose $\mathcal{N}$ at them into smaller components. The *component tree* of $\mathcal{N}$ is a tree whose nodes consists of all those smaller components. The tree is constructed as follows:

> Start with an arbitrary component. While there are still components left out of the tree, choose one such component, say $C$, which contains an NMC separator that is also contained in one or more components in the tree. Add $C$ to the tree by connect it to one of those components that also contain the NMC separator. [3].

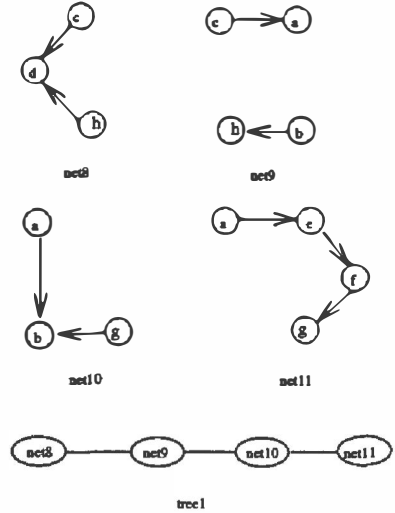

Figure 4: A (the) component tree for net1.

For the Bayesian net net1 in Figure 1, there are three NMC separators are: $\{c, h\}$, $\{a, b\}$, and $\{a, g\}$. As shown in Figure 4 the resulting components are net8 with prior probability $P(c)$ and conditional probabilities $P(d|c, h)$; net9 with conditional probabilities $P(a|c)$ and $P(h|b)$; net10 with conditional probability $P(b|a, g)$; and net11 with conditional probabilities $P(e|a)$, $P(f|e)$, and $P(g|f)$. In this example, there is only one component tree, which is denoted by tree1 and is shown in Figure 4.

The union of two semi-Bayesian nets $(V_1, A_1, R_1, \mathcal{P}_1)$ and $(V_2, A_2, R_2, \mathcal{P}_2)$ is the semi-Bayesian net $(V_1 \cup V_2, A_1 \cup A_2, R, \mathcal{P}_1 \cup \mathcal{P}_2)$, where $R$ is the set of unspecified roots of $(V_1 \cup V_2, A_1 \cup A_2, \mathcal{P}_1 \cup \mathcal{P}_2)$.

Suppose $\mathcal{T}$ is a component tree of a semi-Bayesian net $\mathcal{N}$, and $\mathcal{N}_1$ is a leaf of $\mathcal{T}$. Then $\mathcal{N}$ decomposes into $\mathcal{N}_1$ and $\mathcal{N}_2$ — the union of all other nodes of $\mathcal{T}$. Let $Q$ be a query in $\mathcal{N}$ and let $Q_1$ be the induced query in $\mathcal{N}_1$. Then we can use the answer to $Q_1$ to update the query $Q$.

---
[3] It can be proved that this construction does result in a tree

## 7  COMPONENT TREE PROPAGATION

We are now ready to give the component tree propagation algorithm. Let $\mathcal{N}$ be a parametric semi-Bayesian net with parameters $W$. Here is our algorithm for computing the answer to the query $P_{\mathcal{N}}(X, Y = Y_0 : W)$.

**Main**$(\mathcal{N}, (X, Y = Y_0))$:

1. If $\mathcal{N}$ has NMC separators, call the procedure **Serial-Reduction**$(\mathcal{N}, (X, Y = Y_0))$.
2. If $\mathcal{N}$ does not have any NMC separators, call the procedure **Parallel-Reduction1**$(\mathcal{N}, (X, Y = Y_0))$.

**Serial-Reduction**$(\mathcal{N}, (X, Y = Y_0))$:

1. Decompose $\mathcal{N}$ at all its NMC separators, and construct a component tree $\mathcal{T}$.
2. **While** there is at least two nodes in $\mathcal{T}$ do
   - Pick a leaf $\mathcal{N}_1$ of $\mathcal{T}$ by calling the procedure **Pick-leaf**$(\mathcal{T}, (X, Y = Y_0))$,
   - Call **Main** to compute the induced subquery $Q_1$ in $\mathcal{N}_1$,
   - Append the answer of $Q_1$ to the component that is the neighbor of $\mathcal{N}_1$ in $\mathcal{T}$.
   - Remove $\mathcal{N}_1$ from $\mathcal{T}$ and use the answer to $Q_1$ to update the current query.
3. **When** there is only one node left in $\mathcal{T}$, call **Main** to compute the current query.

In each pass of the while loop, the current query is updated. According to Theorem 3, the answer to the current query is the same as the answer to the updated query. Thus, the answer to the current query when there is only one node left in $\mathcal{T}$ is the same as the answer to the original query $P_{\mathcal{N}}(X, Y = Y_0 : W)$.

The input of to the procedure **Parallel-Reduction1** is parametric semi-Bayesiant net $\mathcal{N}$ and a query $P_{\mathcal{N}}(X, Y = Y_0 : W)$ such that $\mathcal{N}$ does not have any NMC separators. The output is the answer to the query.

**Parallel-Reduction1**$(\mathcal{N}, (X, Y = Y_0))$:

If $\mathcal{N}$ is simple, then compute the answer to the query $P_{\mathcal{N}}(X, Y = Y_0 : W)$ directly by marginalization,
else

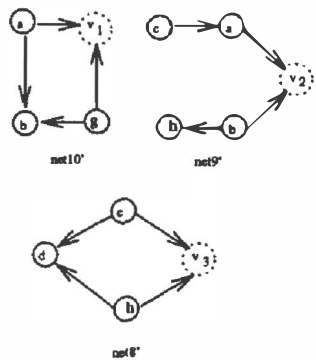

Figure 5: Semi-Bayesian nets created in computing $P_{net1}(d, e)$.

1. Pick a laden node $l$ by calling the procedure **Pick-laden-node**$(\mathcal{N}, (X, Y = Y_0))$,
2. Decompose $\mathcal{N}$ at the set $\pi(l)$ of parents of $l$ into $\mathcal{N}_1$ and $\mathcal{N}_2$,
3. Call **Main** to compute the induced subqueries in $\mathcal{N}_1$ and in $\mathcal{N}_2$,
4. Combine the answers to the subqueries using equation (6).

Our primary investigations indicate that the procedures **Pick-leaf** and **Pick-laden-node** are important to the the performance of the algorithm. How can one define those procedures so that the algorithm achieves its optimal performance and how does this optimal performance compare to the performance of the clique tree propagation approach are topics for future research.

To end this section, let us look at an example. Consider computing $P_{net1}(d, e)$ in the Bayesian net net1 shown in Figure 1. Since net1 has NMC separators, the procedure **Serial-Reduction** will be called. The procedure will first decompose net1 into components (semi-Bayesian nets) net8, net9, net10, and net11 (Figure 4). It will then arrange those components into a component tree tree1, which has two leaves net8 and net11. If the procedure **Pick-leaf** first returns net11, then the induced query $P_{net11}(e, a, g)$ will first be computed. The answer will be appended to the neighbor net10 of net11 in tree1, resulting in the parametric semi-Bayesian net net10' (Figure 5) with auxiliary variable $v_1$ and parameter $e$. The answer will also be used to update the query $P_{net1}(d, e)$ to the new query $P_{net12}(d, v_1 = 0 : e)$, where net12 stands for the union of net8, net9, and net10'.

After net11 is removed from tree1, there are again two leaves net8 and net10'. The current query $P_{net12}(d, v_1 = 0 : e)$ induces the query $P_{net10'}(a, b, v_1 = 0 : e)$ in net10'. If net10' is chosen by **Pick-leaf** this time, the induced query will be computed. Its answer will be appended to net9, resulting in net9' (Figure 5). The answer will also be used to update the current query $P_{net12}(d, v_1 = 0 : e)$ to the new query $P_{net13}(d, v_2 = 0 : e)$, where net13 stands for the union of net8 and net9'.

If **Pick-leaf** now picks net9', then the query $P_{net9'}(c, h, v_2 = 0 : e)$ induced in net9' by the current query $P_{net13}(d, v_2 = 0, : e)$ will be computed. Its answer will be appended to net8, resulting in net8'. And the answer will also to used to update the current query $P_{net13}(d, v_2 = 0 : e)$ to the new query $P_{net8'}(d, v_3 = 0 : e)$.

On the next level, $P_{net10'}(a, b, v_1 = 0 : e)$ and $P_{net8'}(d, v_3 = 0 : e)$ will be computed by the procedure **Parallel-Reduction1** since the parametric semi-Bayesian nets net10' and net8' do not have any NMC separators. On the other hand, net11 and net9' do have NMC separators. So, the procedure **Serial-Reduction** will again be called in computing both $P_{net11}(a, g, e)$ and $P_{net9'}(c, h, v_2 = 0 : e)$.

We notice that there was no triangulation in the above process of computing $P_{net1}(d, e)$, while triangulation is a must in the clique tree propagation approach (see section 2).

## 8 CONCLUSIONS

In this paper, we have described a approach for computing posterior probabilities in a Bayesian net, which sidesteps the triangulation problem. Our approach begin by decomposing the Bayesian net into smaller components by making use of Tarjan's algorithm for decomposing undirected graphs at all their minimal complete separators. Like the clique tree approach, our approach arranges those components into a tree, and then computes posterior probabilities by appropriately passing information around in that tree. Unlike the clique tree approach, the computation in each component is not carried out by direct marginalization. Rather it is carried by repeating the whole procedure from the beginning. Thus, the need for triangulation is avoided.

How does the performance of our approach compare to that of the clique tree propagation approach based on an optimal triangulation? This question is yet to be answered. Our hope is that proper choices of the procedure **Pick-leaf** and **Pick-laden-node** could ensure the performance of our approach lay close to optimal.

## Acknowledgement

The first author gained his understanding of Bayesian nets when he was with Glenn Shafer and Prakash Shenoy at Business School, University of Kansas from October 1987 to October 1988 and from September 1989 to August 1990. The paper has benefited from comments by D'Ambrosio and the reviewers for UAI 92. Research is supported by NSERC Grant OG-POO44121.